\documentclass[12pt,a4paper]{article}

\usepackage{fontspec}
\IfFontExistsTF{Liberation Serif}{\setmainfont[Ligatures=TeX]{Liberation Serif}}{\setmainfont{cmunrm.otf}[BoldFont=cmunbx.otf,ItalicFont=cmunti.otf,BoldItalicFont=cmunbi.otf]}
\usepackage[margin=25mm]{geometry}
\usepackage{pgfplots}
\pgfplotsset{compat=1.17}
\usepackage[english]{babel}
\usepackage{hyperref}
\usepackage{amsmath}
\usepackage{amssymb}
\usepackage{booktabs}
\usepackage{array}
\usepackage{tabularx}
\usepackage{multirow}
\usepackage{graphicx}

\newcolumntype{P}[1]{>{\small\raggedright\arraybackslash}p{#1}}

\begin{document}

\title{On the Applicability of $p$-adic Numbers in Computational Modules for Neural Networks}

\author{%
  Vladimir Evgrafov%
  \thanks{Vladimir Evgrafov, Lomonosov Moscow State University, Faculty of Computational Mathematics and Cybernetics (CMC), Moscow, Russia.}
}

\date{}
\maketitle

\begin{abstract}
The applicability of $p$-adic arithmetic as an intermediate computational module in an otherwise unchanged machine-learning pipeline is investigated. A one-feature linear regression on the scikit-learn Diabetes dataset serves as the test model. The fitted coefficients are represented in $\mathbb{Q}_p$, predictions are computed in that field, and rational reconstruction is applied to the outputs. In the pyAdic experiments, rational reconstruction either returns no result or produces values that differ from the baseline predictions; changing computational precision can change both the magnitude and sign of the reconstructed value. In SageMath, forward conversions succeed, but the reconstructed predictions differ from the baseline. An alternative representation based on digit reversal fails to preserve ordinary addition because of carry propagation. Under the tested settings, none of the test predictions is reconstructed correctly. These results show that the tested conversion and reconstruction procedures do not provide a compatible intermediate module even for linear regression. Models defined directly over $p$-adic data and operations offer a direction for further investigation.
\end{abstract}

\par

%%--------------------------------------------------------------------
\section{Introduction}
%%--------------------------------------------------------------------

Kurt Hensel introduced $p$-adic numbers at the end of the nineteenth century for the study of integers and equations in number fields. They are used in number theory and algebra and provide a metric and topology distinct from those of the real numbers~\cite{r13,r14,r15}.

Formally, the field $\mathbb{Q}_p$ is the completion of $\mathbb{Q}$ with respect to the $p$-adic norm $\|\cdot\|_p$. Let $p$ be a prime. For a rational number $a$ the norm is
\[
\|a\|_p =
\begin{cases}
0, & a = 0, \\
p^{-k}, & a \neq 0,\ a = p^k \frac{m}{n},\ p \nmid m,\ p \nmid n,
\end{cases}
\]
that is, $k$ is the exponent of $p$ in $a$. Classically, a $p$-adic number is a sequence $x = \{x_1, x_2, \ldots\}$ in which $x_n$ is a residue modulo $p^n$. Equivalently, every element of $\mathbb{Q}_p$ can be written as a series
\[
x = \sum_{i=k}^{\infty} a_i p^i, \qquad k \in \mathbb{Z},\ a_i \in \{0, 1, \ldots, p-1\}.
\]
In the $p$-adic metric, proximity is determined by agreement in the low-order digits of the $p$-adic expansion, in contrast to the role of leading digits in real positional representations.

Applications of $p$-adic methods include Diophantine equations, congruences, algebraic geometry, dynamical systems~\cite{r4}, and mathematical physics. The review by Dragovich et al.~\cite{r16} surveys three decades of $p$-adic mathematical physics, including applications to string theory and biological models.

Computer implementations of $p$-adic arithmetic require finite representations or procedures for generating digits, together with a method of managing precision. Caruso's lecture notes~\cite{r2} compare the principal computational paradigms. Caruso, Roe and Vaccon~\cite{r18} examine precision tracking, while van der Hoeven~\cite{r19} develops lazy and relaxed computation with power series and $p$-adic numbers. Implementations are available in computer algebra systems, including SageMath~\cite{r9}, and in standalone libraries, including pyAdic~\cite{r7}.

In machine learning, Zambrano-Luna and Z\'u\~niga-Galindo developed $p$-adic cellular neural networks and considered their application to image processing~\cite{r11,r12}. Z\'u\~niga-Galindo, He and Zambrano-Luna related $p$-adic statistical field theory to deep Boltzmann machines; their construction yields convolutional networks with $O(N)$ rather than $O(N^2)$ parameters~\cite{r20}.

The present study concerns a different integration scheme: a module that accepts rational or floating-point inputs, computes in $\mathbb{Q}_p$, and returns rational outputs to an existing model. The intended benefit is reuse of the surrounding architecture, datasets, and software without redefining the entire pipeline over a $p$-adic field. A further motivation is the potential application of such a module to the exploding-gradient problem.

The exploding-gradient problem arises when errors are backpropagated through many layers or time steps of a recurrent network. By the chain rule, the derivative involves a product of the Jacobian matrices of the individual transformations. Let $g_L$ denote the loss gradient with respect to the final layer output and $J_\ell$ the Jacobian of layer $\ell$. Then
\[
    g_0 = J_1^{\mathsf T} J_2^{\mathsf T}\cdots J_L^{\mathsf T}g_L.
\]
In real arithmetic, this product can amplify some directions as network depth increases. Large gradients can cause excessively large parameter updates, unstable training and, in extreme cases, numerical overflow. Gradient norm clipping is a common way to limit this effect~\cite{r21}.

The motivation for a $p$-adic approach is a different measure of the size of intermediate values. The $p$-adic norm satisfies the strong triangle inequality
\[
    \|u+v\|_p\leq\max\{\|u\|_p,\|v\|_p\}.
\]
Thus, addition alone cannot increase the norm beyond the largest norm of its summands. Define the vector norm $\|v\|_{p,\infty}=\max_i\|v_i\|_p$ and the matrix norm $\|A\|_{p,\max}=\max_{i,j}\|a_{ij}\|_p$. Then
\[
    \|g_0\|_{p,\infty}
    \leq \left(\prod_{\ell=1}^{L}\|J_\ell\|_{p,\max}\right)
    \|g_L\|_{p,\infty}.
\]
If all Jacobian entries in a model over $\mathbb{Q}_p$ belong to $\mathbb{Z}_p$, every factor on the right is at most one: backpropagation does not increase the $p$-adic gradient norm. For example, repeated multiplication by $p$ produces the factor $p^L$, whose real magnitude grows while its $p$-adic norm $\|p^L\|_p=p^{-L}$ decreases. This illustrates a possible mechanism for controlling growth in the $p$-adic metric.

Such control requires restrictions on derivatives and does not follow merely from changing the numerical type. Conversely, the factor $1/p$ has norm $p$ and causes the $p$-adic norm to grow under repeated multiplication. Boundedness in the $p$-adic metric also does not imply bounded reconstructed real values: in the example above, $p^L$ remains large as a real number. Applying this mechanism to training therefore requires differentiable $p$-adic transformations, a loss function and a parameter-update rule, together with correct reconstruction when returning to a real-valued model. Here this possibility serves as motivation; the linear-regression experiments test arithmetic-module compatibility and do not demonstrate mitigation of exploding gradients.

The experiments in this study examined this scheme using linear regression and yielded a negative result. The problem statement is given in Section~\ref{sec:problem}, the computational tools in Section~\ref{sec:context}, and the experimental design and results in Sections~\ref{sec:design} and~\ref{sec:results}. Section~\ref{sec:verification} presents a quantitative assessment of reconstruction. Sources of error, further research directions, and the conclusions are discussed in Sections~\ref{sec:issues}--\ref{sec:conclusion}.

%%--------------------------------------------------------------------
\section{Problem Statement}\label{sec:problem}
%%--------------------------------------------------------------------

The proposed module receives rational or floating-point inputs, represents them in $\mathbb{Q}_p$, performs arithmetic in that field, and reconstructs a rational output. Conversion is confined to the module boundary; the stored training data and the surrounding model retain their original representation. The module is intended to preserve compatibility with the existing pipeline.

The scheme requires both forward conversion and rational reconstruction. A rational input can be represented in $\mathbb{Q}_p$ to a specified precision. Reconstruction from a finite $p$-adic approximation is not uniquely determined without additional restrictions, since multiple rational numbers can agree with the same approximation.

To map $x = \frac{r}{s} \in \mathbb{Q}$ into $\mathbb{Z}/p^k\mathbb{Z}$, it is enough to find the inverse $q = s^{-1}$ of the denominator, i.e. an element with $q\cdot s \equiv 1 \pmod{p^k}$. When $\gcd(s, p) = 1$, this is done with the extended Euclidean algorithm or with continued fractions.

In the general case, powers of $p$ are first separated from the numerator and denominator:
\begin{equation}
    x = \frac{r}{s}\cdot p^v.
\end{equation}
If the denominator still has a common factor with $p$, $\gcd(s, p) = c \neq 1$, the following transformation is applied:
\begin{equation}
    x = \frac{p^{v'}}{p^{v'}} \cdot \frac{c^{v'}}{c^{v'}} \cdot \frac{r}{s} \cdot p^{v} = \frac{r\cdot (p/c)^{v'}}{s / c^{v'}}\cdot p^{(v-v')}.
\end{equation}
The remaining fraction is a unit of $\mathbb{Z}_p$.

Linear regression is used as a minimal test of the conversion procedure within a model computation. Agreement with its baseline predictions is a necessary requirement before applying the same procedure to more complex neural-network models.

%%--------------------------------------------------------------------
\section{Tools and Implementation Paradigms}\label{sec:context}
%%--------------------------------------------------------------------

The representation of $p$-adic numbers and the treatment of precision are relevant to the interpretation of library outputs. Caruso~\cite{r2} distinguishes three principal computational approaches.

\textit{Zealous arithmetic} tracks precision explicitly and is a $p$-adic analogue of interval arithmetic. Each value is represented together with its precision as a ball
\[
I_{N,a} = \{x \in \mathbb{Q}_p : |x - a| \leq p^{-N}\} = a + O(p^N),
\]
and each operation updates both the computed digits and the precision bound. This finite representation allows precision losses to be tracked during computation.

\textit{Lazy and relaxed computations}~\cite{r19} represent a number by a procedure that generates digits as required. Lazy methods defer the computation of digits until they are requested. Relaxed methods reuse previously computed digits and progressively extend results when additional precision is required, rather than recomputing the entire approximation.

\textit{$p$-adic floating-point arithmetic} represents a number by a fixed number of significant $p$-adic digits and an exponent. The number of stored digits is fixed, but this representation alone does not provide a certified precision bound for the result of every operation.

The choice of representation depends on the computational task. The experiments use the Python library pyAdic~\cite{r7} and SageMath~\cite{r9}. SageMath provides fixed modulus rings and implementations with capped absolute and capped relative precision, offering different approaches to precision management. The baseline regression is implemented in scikit-learn~\cite{r6}.

The $p$-adic cellular networks~\cite{r11,r12} are not used as experimental baselines because they address different model constructions. The observations concern only the tested intermediate arithmetic module and do not constitute an assessment of $p$-adic methods in neural networks generally.

%%--------------------------------------------------------------------
\section{Experimental Design}\label{sec:design}
%%--------------------------------------------------------------------

The baseline experiment uses the Diabetes dataset from scikit-learn~\cite{r6}. A one-feature linear regression is fitted using the original floating-point representation. Its slope and intercept are converted to $p$-adic numbers, predictions are computed in $\mathbb{Q}_p$, and rational reconstruction is applied to the results.

The comparison evaluates whether the sequence of input conversion, arithmetic, and rational reconstruction reproduces the baseline predictions. This tests the compatibility of the $p$-adic computation with the input and output representations of the model.

Additional experiments examine sensitivity to $p$-adic precision, an analogous computation in SageMath, and an alternative representation based on reversal of the digit order.

%%--------------------------------------------------------------------
\section{Results}\label{sec:results}
%%--------------------------------------------------------------------

\subsection{Linear Regression with $p$-adic Coefficients}

The baseline regression metrics are given in Section~\ref{sec:verification}. In the experiment, the coefficients $a$ and $b$ were converted to $p$-adic representations $a_p$ and $b_p$ and used to evaluate the corresponding linear expression. The reconstructed outputs differed from the baseline predictions~\cite{r0}.

Most reconstructions in the displayed output returned \texttt{False} rather than a rational value. Thus, the tested reconstruction procedure failed to recover predictions compatible with the baseline model.

A reduced test isolated the conversion of a single floating-point input to a $p$-adic representation and back. The value $0.04716281$ was converted with $p = 2$ and absolute precision 40, followed by the built-in \texttt{rationalise} function. The reconstructed value had the opposite sign and differed by an order of magnitude~\cite{r0}. The conversion discrepancy therefore occurred even without a regression calculation.

\subsection{Dependence on Precision}

Changes in the reconstructed values with the precision setting motivated a separate examination of precision dependence.

In pyAdic, the input is first represented as a fraction $\frac{a}{b}$ of integers using Python's \texttt{fractions} module. The $p$-adic representations $a_p$ and $b_p' = (\frac{1}{b})_p$ are then computed and multiplied: $a_p \cdot b_p'$.

The reconstructed values varied substantially with precision: the same rational input produced different outputs, sometimes with opposite signs. A floating-point value such as $0.04716281$ becomes an exact fraction with large numerator and denominator values, making rational reconstruction sensitive to the available precision. Under the tested settings, pyAdic did not provide stable recovery of the rational values required by the intermediate module.

\subsection{SageMath}

An analogous scenario was examined in SageMath~\cite{r9} to assess whether discrepancies also occurred with another implementation. Its fixed modulus rings and implementations with capped absolute and capped relative precision provide alternative ways of managing the finite representations discussed in~\cite{r2}.

In SageMath, capped relative precision was set to 80 significant $p$-adic digits. With this setting, rational-to-$p$-adic conversions completed successfully, unlike the problematic pyAdic cases. However, rational reconstruction of the computed results did not reproduce the baseline predictions~\cite{r0}.

The reconstruction discrepancy therefore also occurred in a second computational environment. Successful forward conversion alone was insufficient to preserve compatibility with the baseline model after the complete round trip.

\subsection{Reversing the Digit Order}

The alternative representation reverses the digit order so that leading digits become trailing digits and conversely. Its motivation is the different role of digit positions in real and $p$-adic metrics. Such a representation does not, by itself, preserve arithmetic operations.

In ordinary positional addition, carries propagate towards higher powers of the base. After digit reversal, preserving the original addition would require propagation in the opposite direction in the reversed record. The resulting discrepancy is illustrated by the relation
\begin{equation}
    \operatorname{int}\bigl(\operatorname{inv}(a) + \operatorname{inv}(b)\bigr) \neq a + b.
\end{equation}

Fig.~\ref{fig:inversion} reproduces the decimal example from the experiments in this study. The right-hand expression uses the reversed digit order with a corresponding shift of the decimal point. The sums differ: reversing the digits of 309579000 gives 0.00975903 rather than 0.10885902. The carry-direction problem also motivates consideration of binary positional arithmetic, including the significands used in IEEE~754 representations.

\begin{figure}[htbp]
  \centering
  \[
\begin{array}{r}182617400\\+\;126961600\\\hline309579000\end{array}
\quad\not\longleftrightarrow\quad
\begin{array}{r}0.04716281\\+\;0.06169621\\\hline0.10885902\end{array}
\]
  \caption{Addition in the original and digit-reversed representations, reproduced from the experimental materials~\cite{r0}. Ordinary addition does not preserve the proposed correspondence in this example because of carry propagation.}
  \label{fig:inversion}
\end{figure}

Preserving arithmetic under this representation requires explicit definitions and implementations of the transformed operations rather than direct reuse of ordinary addition. This conflicts with the aim of inserting a module without changing the surrounding computational system. Errors in low-order digits of the reversed representation are especially relevant in the $p$-adic metric, which also complicates the intended application to exploding gradients.

\begin{table}[htbp]
  \caption{Summary of the experiments}
  \label{tab:summary}
  \centering
  \small
  \begin{tabular}{P{2.7cm}P{3.2cm}P{2.0cm}}
    \toprule
    \textbf{Approach} & \textbf{Experimental observation} & \textbf{Consequence} \\
    \midrule
    pyAdic + linear regression & Most displayed reconstructions return \texttt{False} & Tested procedure does not reproduce baseline \\
    Varying precision & Reconstructed values vary, including sign changes & Unstable recovery under tested settings \\
    SageMath & Forward conversions complete; reconstructed values differ & Round trip does not preserve predictions \\
    Reversed digits & Ordinary addition does not preserve the representation & Transformed operations required \\
    \bottomrule
  \end{tabular}
\end{table}

%%--------------------------------------------------------------------
\section{Quantitative Assessment of Reconstruction}\label{sec:verification}
%%--------------------------------------------------------------------

The main regression experiment uses the Diabetes dataset from scikit-learn~\cite{r6}, feature index 2, the last 20 observations as the test set, and pyAdic~\cite{r7} with $p=2$ and precision 50. Agreement after reconstruction is evaluated with an absolute tolerance of $10^{-6}$.

The baseline regression gives a coefficient of 938.23786125, a mean squared error of 2548.07, and $R^2 = 0.47$. None of the test predictions is recovered correctly after the $p$-adic round trip: the reconstruction outcome is 0 out of 20.

\begin{figure*}[htbp]
  \centering
  \begin{tikzpicture}
\begin{axis}[width=.48\linewidth,height=6cm,title={Baseline regression},xlabel={BMI (normalized)},ylabel={Disease progression},tick label style={font=\scriptsize},label style={font=\small}]
\addplot[only marks,mark=*,mark size=1.5pt] coordinates {(0.077863388,233.0) (-0.039618128,91.0) (0.011039039,111.0) (-0.040695940,152.0) (-0.034229068,120.0) (0.005649979,67.0) (0.088641508,310.0) (-0.033151256,94.0) (-0.056863122,183.0) (-0.030995632,66.0) (0.055229334,173.0) (-0.060096558,72.0) (0.001338730,49.0) (-0.023450947,64.0) (-0.074108115,48.0) (0.019661536,178.0) (-0.015906263,104.0) (-0.015906263,132.0) (0.039062153,220.0) (-0.073030303,57.0)};
\addplot[blue,domain=-.08:.10] {938.23786125*x+152.91886183};
\end{axis}
\end{tikzpicture}\hfill
\begin{tikzpicture}
\begin{axis}[width=.48\linewidth,height=6cm,title={Reconstruction: 0 of 20},xlabel={Test point index},xmin=0,xmax=21,ymin=-.4,ymax=2.4,ytick={0,1,2},yticklabels={No number,Outside tolerance,Within tolerance},tick label style={font=\scriptsize},label style={font=\small},xtick={1,5,10,15,20}]
\addplot[only marks,mark=x,red] coordinates {(1,0) (2,0) (3,0) (4,0) (5,0) (6,0) (7,0) (8,0) (9,0) (10,0) (11,0) (12,0) (13,0) (14,0) (15,0) (16,0) (17,0) (18,0) (19,0) (20,0)};
\end{axis}
\end{tikzpicture}
  \caption{Experimental results. Left: the Diabetes test split and baseline regression. Right: reconstruction status (pyAdic, $p=2$, precision 50; 20 test points). Successful reconstruction is assessed using an absolute tolerance of $10^{-6}$.}
  \label{fig:verification}
\end{figure*}
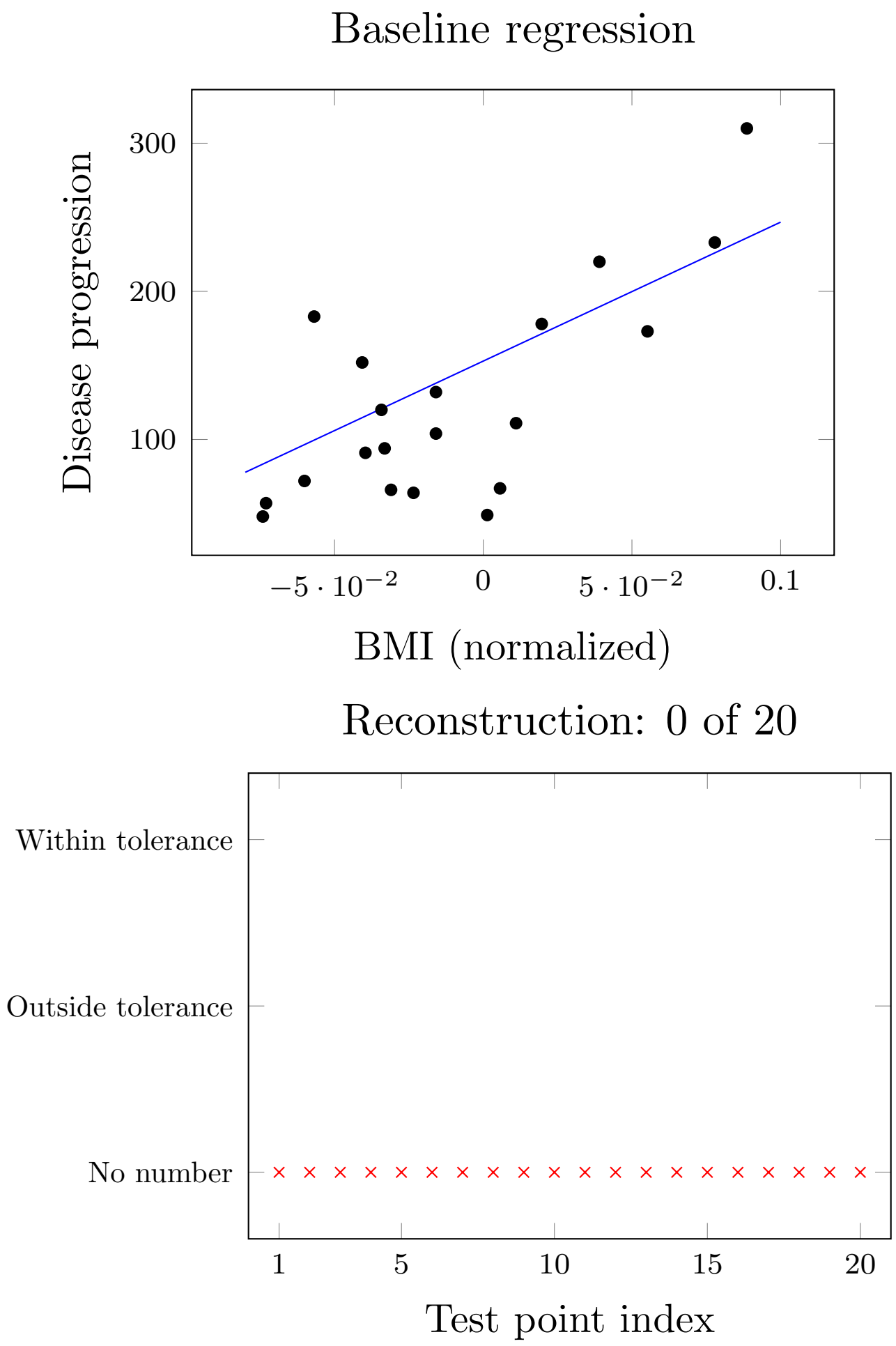

Fig.~\ref{fig:verification} illustrates the regression and reconstruction results. The left panel depicts the test data and fitted line. The right panel shows the reconstruction status of each test point.

Rational reconstruction fails to recover the test predictions for the specified regression setup and parameter choices.

%%--------------------------------------------------------------------
\section{Possible Problems and Sources of Error}\label{sec:issues}
%%--------------------------------------------------------------------

The experiments reveal sensitivity to precision and to small input changes. Input representation, available precision, and rational reconstruction are therefore central to the reliability of the intermediate module. The different real and $p$-adic metrics make preservation of the baseline model's predictions a substantive requirement rather than an automatic consequence of changing the number representation.

For digit reversal, the carry mechanism prevents direct reuse of ordinary addition. Transformed operations would be needed to maintain the intended correspondence, contrary to the requirement of minimal changes to the existing system.

One possible explanation of the pyAdic results is incomplete handling of special cases in conversion, such as factors $p^n$ in a rational denominator. The observed failures motivate closer analysis of these cases and of the precision needed for rational reconstruction.

%%--------------------------------------------------------------------
\section{Next Steps}\label{sec:next}
%%--------------------------------------------------------------------

Further investigation may consider datasets defined directly over $p$-adic fields. Such a formulation avoids the rational-to-$p$-adic round trip that produced reconstruction errors in the experiments.

Another direction is automatic selection of the prime $p$ and the computational precision under an explicitly defined reconstruction or prediction criterion.

Architectures formulated over $p$-adic numbers provide a further research direction. Cellular neural networks use local connections between neighboring neurons and have applications in parallel computation and image processing. In~\cite{r11,r12}, $p$-adic representations encode neuron states, with digits corresponding to individual neurons. Further work includes specifying the architecture, training and weight-update rules, and comparison with relevant baselines.

%%--------------------------------------------------------------------
\section{Discussion}\label{sec:discussion}
%%--------------------------------------------------------------------

The experiments demonstrate that the tested conversion--computation--reconstruction pipeline does not preserve the predictions of even a linear regression. Converting the fitted coefficients to $p$-adic numbers is insufficient to obtain a compatible intermediate computational module.

This result concerns the tested implementations and parameter choices, not $p$-adic machine-learning methods in general. The practical difficulty lies in recovering the rational outputs expected by the surrounding model.

Two aspects of the difficulty are distinct. Reconstruction failures and precision-dependent outputs concern the numerical behavior of the tested procedures. Digit reversal raises a representation-level problem: preserving addition requires changing the carry mechanism. Both aspects conflict with the aim of retaining the surrounding computational system unchanged.

The negative result identifies a limitation of direct substitution of the number representation in an existing model. A $p$-adic model instead requires a formulation in which the data and operations are $p$-adic from the outset, or an intermediate module explicitly designed to preserve the required input--output correspondence.

%%--------------------------------------------------------------------
\section{Conclusion}\label{sec:conclusion}
%%--------------------------------------------------------------------

The present study examined an intermediate $p$-adic computational module motivated by integration into existing neural-network systems. The actual test model was a one-feature linear regression: its fitted coefficients and inputs were represented in $\mathbb{Q}_p$, and its computed outputs were subjected to rational reconstruction.

The experiments yield unsuccessful and precision-dependent reconstructions in pyAdic, discrepancies after reconstruction in SageMath, and a digit-reversal example that does not preserve ordinary addition. Under the tested settings, 0 of the 20 test predictions are reconstructed successfully.

The tested procedures therefore do not provide a compatible intermediate $p$-adic module for the baseline model. Further research includes models based on $p$-adic data and operations, such as $p$-adic cellular neural networks~\cite{r11,r12} and related constructions~\cite{r20}, automatic selection of $p$ and precision, and architectures designed around alternative numerical representations.

\end{document}